# A single scale retinex based method for palm vein extraction

Chongyang WANG[1,2], Ming PENG[1,2], Lingfeng XU[1,2], Tong CHEN[1,2,*]

[1]School of Electronic and Information Engineering
Southwest University
Chongqing, China
*Email: c_tong @swu.edu.cn

[2]Chongqing Key Laboratory of Nonlinear Circuit and
Intelligent Information Processing
Southwest University
Chongqing, China

*Abstract*—Palm vein recognition is a novel biometric identification technology. But how to gain a better vein extraction result from the raw palm image is still a challenging problem, especially when the raw data collection has the problem of asymmetric illumination. This paper proposes a method based on single scale Retinex algorithm to extract palm vein image when strong shadow presents due to asymmetric illumination and uneven geometry of the palm. We test our method on a multispectral palm image. The experimental result shows that the proposed method is robust to the influence of illumination angle and shadow. Compared to the traditional extraction methods, the proposed method can obtain palm vein lines with better visualization performance (the contrast ratio increases by 18.4%, entropy increases by 1.07%, and definition increases by 18.8%).

## I. INTRODUCTION

Palm vein recognition is a novel and promising biometric identification technology [1]. It has received more attention recently, because the palm vein pattern is intrinsic, unchangeable, and easily captured by ordinary sensors. Many factors, especially asymmetric illumination, can affect the quality of original palm image. The low quality of palm image will largely increase the difficulty to extract vein pattern. In order to obtain better vein pattern extraction result, various methods of image enhancement are used before the pattern extraction process.

In this paper, we proposed a method to enhance the palm vein image and visualize the vein pattern under strong shadow effect. First, we use hyperspectral imager to acquire palm image data in wavebands near the 850nm wavelength. Secondly, we use images from different bands to create a normalized image with an average wavelength of 850nm. Thirdly, we apply Retinex [2] method to reduce the influences caused by shadow. Additionally, we stretch the dynamic range and smoothen the processed image by using histogram equalization [3] and median filter. Finally, we set specific threshold value to acquire binaryzation image [4] of palm vein and remove the unnecessary part of vein.

Compared with other vein image enhancement methods [5]-[7], the proposed method can successfully reduce the influences caused by shadow while keep the palm vein pattern unchanged, and eventually produce better visualization result of palm vein.

## II. BACKGROUND

The shadow on the palm caused by the asymmetric illumination environment or uneven geometry of the palm will increase the difficulty to extract the palm vein pattern. Under the circumstance that there is no definite symmetric illumination environment, we normally apply image enhancement method to reduce influence caused by shadow.

Various traditional image enhancement algorithms have been applied to the research area of hand vein image processing. Rossan et al. [5] used contrast limited adaptive histogram equalization method (CLAHE) to increase the contrast of palm vein image. However, this method didn't work well when shadow existed. Kang W et al. [6] applied Difference of Gaussian-Histogram Equalization (DoG-HE) to gain palm vein pattern image, but the shadow on the palm can largely influence the result. You Lin et al. [7] used Gaussian Low Pass Filter to enhance the finger vein image. However, the final image was vague, and could be easily affected by asymmetric illumination. To sum up, the aforementioned methods failed to perform well in the asymmetric illumination environment (see section III C for results).

The Retinex theory introduced by Land [2] divides a two dimensional image into two components: illumination component and reflection component. A common use of this theory is to estimate the illumination component of an image, overcome the bad influence caused by asymmetric illumination environment through processing it, and retain features of reflection component. The Retinex method can successfully reduce the influence caused by asymmetric illumination with more information of the image reserved. The method currently has a wide use in medical image enhancement area, such as using Retinex to enhance the image contrast in magnetic resonance imaging (MRI) [8]. Also it can solve the low contrast problem caused by complex illumination environments in RGB images [9] and improve the visibility of images taken under foggy weather [10]. In this paper, we employ Retinex method to solve the problem of shadow in palm vein image, while also retain the vein pattern.

Multispectral imaging technology is widely used in palm vein imaging area, which can acquire image in a wavelength ranging from visible light to near-infrared light and produce images in 100 to 10000 continuous wavebands. Multispectral imaging can obtain more information of the object than traditional imaging technologies. In acquiring the palm vein image, it can provide better results in specific band [11]. Multispectral imaging can also be applied to image sensing area, since it can display the infrared image of human body [12].

The typical wavelength of light source in palm vein identification system is 760nm, 850nm, 890nm and 940nm. For pattern extraction and feature matching, 850nm wavelength can give the best results [11].

### III. PALM VEIN IMAGE EXTRACTION

#### A. Instrument and Experiment setup

The imaging instrument used in our experiment is a multispectral imaging system, which consists of a spectrometer (SPECIM V10E, Finland) and a photoelectric sensor (CCD, Lumenera Infinity3-1M, Canada). The image cube was produced by a scanner whose motion trail and the trigger process were all automatically synchronized by a computer. The light source in our experiment was two halogen lights that shines from the top directly to a palm in 45° angle. During the image capturing process, the palm was kept clean and open steadily, the distance between the imager and the palm was 1m. The palm was approximately parallel to the lens of the imager. Shadows are produced on the center part of the palm due to the illumination angle and geometry of the palm (see Fig 1).

#### B. Palm vein image generating and normalization

Palm image acquired in 850nm wavelength can give the best definition and recognition performance [11]. Therefore, we choose the cube image with average wavelength of 850nm and bandwidth of 10nm to create a two dimensional palm image by averaging images at every sub waveband. The palm image f(x, y) is then normalized by using

$$I(x,y) = \frac{f(x,y) - \min}{\max - \min} \quad (1),$$

where $f(x, y)$ is the input image, $\max = \max(f(x,y))$, $\min = \min(f(x,y))$, and $I(x, y)$ is the output image after doing gray value normalization, which is shown in Fig. 1 (a).

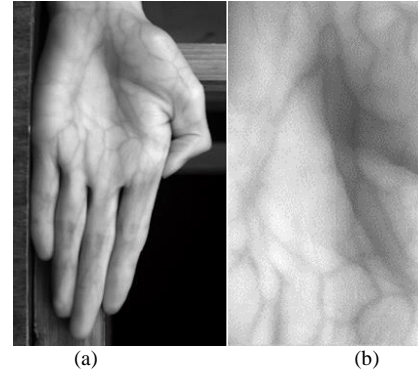

(a)      (b)

Fig. 1. (a) Palm image after doing gray value normalization, (b) Palm image after doing scale normalization

In fig. 1 (a), the palm image included fingers and background, which will increase the processing time and bring in unnecessary parts. Thus, we extract the part containing the palm vein pattern from the original image. The original image has 1564x969 pixels, and we segment out the central part of the palm, which has 360x657 pixels. The central part is shown in Fig. 1 (b).

#### C. Reduce influence caused by asymmetric illumination using Retinex

The Retinex theory introduced by Land [5] divides a two dimensional image into two components, i.e. illumination component and reflection component. The theory can be described by the equation below:

$$S(x,y) = R(x,y) \bullet L(x,y) \quad (2)$$

Where S(x, y) is the input image, R(x, y) is the reflection component, L(x, y) is the illumination component. Usually we need to find methods to estimate the illumination component of an image, and overcome the bad influences caused by asymmetric illumination by processing the illumination component. The image will normally get transferred to log-domain, which can be defined as follows:

$$s = \log S \quad (3),$$

$$l = \log L \quad (4),$$

$$r = \log R \quad (5).$$

Therefore s = l + r.

The key of Retinex algorithm is applying specific mathematic method to estimate the illumination component of an image, and restore the Reflection component by processing the illumination component. Under the assumption that the incident light is smooth, we can use Gaussian function to estimate the illumination component by convolving with the image.

Based on the research by Land [2], the single scale Retinex algorithm proposed by Jobson [13] can be defined as follows:

$$R(x, y) = \log I(x, y) - \log[G(x, y) * I(x, y)] \quad (6),$$

$$G(x, y) = \frac{1}{\sqrt{2\pi\delta}} e^{-\frac{x^2+y^2}{2\delta^2}} \quad (7),$$

where * denotes convolution, I(x, y) is the input image, G(x, y) is the Gaussian function, R(x, y) is the estimated reflection component. $\delta$ is the control neighborhood-wide scale constant, the larger the $\delta$ is, the more obviously the dynamic range is compressed. The smaller the $\delta$ is, the more obviously the contrast is enhanced. In our experiment, the is set to 25.

The result after using Retinex on the central part of the palm (Fig. 2 (a)) is shown in Fig. 2 (e). Since the dynamic range is compressed, we use histogram equalization (HE) [3] to stretch the dynamic range, the result is shown in Fig. 2 (f). We then use median filer to smoothen the image after HE, the result is shown in Fig. 2 (g). Finally, we get the binaryzation image by setting threshold value [4], the result is shown in Fig. 3 (a).

The raw image is also processed by using CLAHE, DoG-HE, and Gaussian low pass filter, the results are illustrated in Fig 2 (b), (c), and (d), respectively. It is seen that these methods are fragile to shadow effect, i.e. the shadow parts on the center of the palm are enhanced by these methods, though the vein are enhanced. However, the proposed method is robust to the shadow effect. It is observed form Fig 2 (f) that only the palm vein and skin line are enhanced and the strong shadow on the palm is removed by using the proposed method.

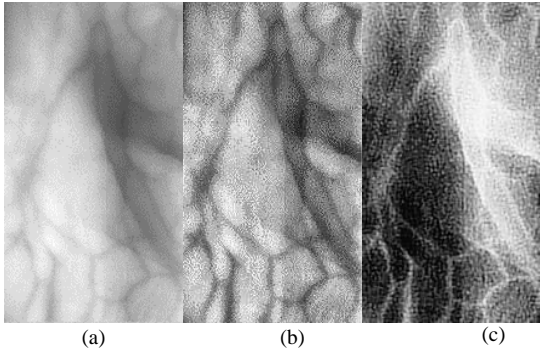

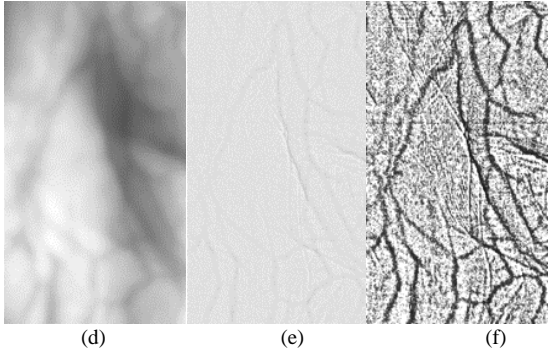

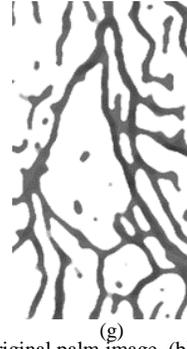

Fig. 2. (a) Original palm image, (b) Palm image after using CLAHE to a, (c) Palm image after using DoG-HE method to a, and the ratio of the two different Gaussian kernels is set to 4:1, (d) Palm image after using Gaussian low pass filter to a, (e) Palm image after using Retinex to a, (f) Palm image after using HE to e, (g) Palm image after using median filter to f

### D. Removing the false image area

As shown in Fig. 3 (a), there exist some isolated image blocks (points) and broken lines near the vein lines. In order to retain useful and continuous vein lines, we adopt morphology method to get rid of the false area. Here, we delete connected domain smaller than 20000 pixels. The result after morphology processing is shown in Fig. 3 (b).

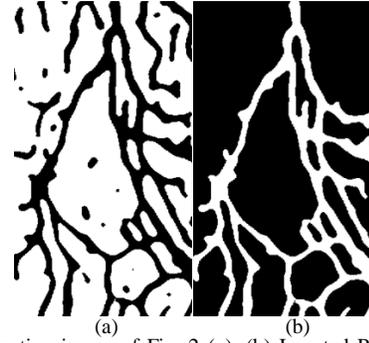

Fig. 3. (a) Binaryzation image of Fig. 2 (g), (b) Inverted Palm image after deleting isolated blocks (points) and broken lines

## IV. COMPARISON

To further demonstrate the efficiency of the proposed method, we compared it with other methods mentioned in this paper, and took contrast, entropy and definition as indicators.

The contrast of an image X (x, y) can be defined as follows:

$$C = \sqrt{\frac{1}{M \times N} \sum_{i=1}^{M} \sum_{j=1}^{N} \left(X(i,j) - \bar{X}\right)^2} \quad (8)$$

$$\bar{X} = \frac{1}{M \times N} \sum_{i=1}^{M} \sum_{j=1}^{N} X(i,j) \quad (9)$$

where C denotes the contrast of an image, the larger the C is, the more obvious the discrimination between vein lines and the background is.

The entropy of an image can be defined as follows:

$$E = -\sum_{i=0}^{L-1} p_i \ln(p_i) \quad (10)$$

where L is the gray scale of image, P denotes the possibility of symmetric gray value. The larger the entropy is, the more information an image is carrying.

The definition D of an image $f(m, n)$ can be defined as follows:

$$D = \frac{1}{MN}\sum_{m=1}^{M}\sum_{n=1}^{N}\sqrt{\left(\Delta I_x(m,n)\right)^2 + \left(\Delta I_y(m,n)\right)^2} \quad (11)$$

where

$$\begin{cases} \Delta I_x(m,n) = f(m,n) - f(m-1,n) \\ \Delta I_y(m,n) = f(m,n) - f(m,n-1) \end{cases} \quad (12)$$

The larger the D is, the more distinct the image is.

And the contrast, entropy and definition of each method are given in table I.

TABLE I.    CONTRAST (C), ENTROPIES (E), AND DEFINITION (D) OF ENHANCED PALM VEIN IMAGES BY USING VARIOUS METHOD.

|  | Indicator | | |
|---|---|---|---|
|  | *Contrast* | *Entropy* | *Definition* |
| Original | 32.8224 | 5.9618 | 2.2416 |
| CLHAE | 53.6735 | 6.9370 | 4.4789 |
| DoG-HE | 64.8192 | 3.3458 | 12.016 |
| Gaussian low pass | 33.6801 | 6.1610 | 2.6822 |
| Proposed | 76.7143 | 7.0119 | 14.2746 |

We can see from table I that our method produces better result of contrast, entropy and definition when processing the same image (Fig. 2 (a)), which indicates that our method is a more effective way to extract the palm vein lines.

To further show the extracted vein lines, we use morphology methods [14] to thining Fig. 3 (b), and the result is shown in Fig. 4. As a consequence, we can see, the proposed method can retain more information of vein pattern and give better visualization result.

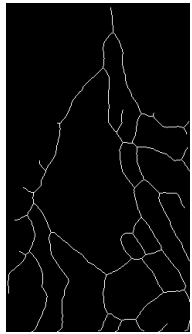

Fig. 4.   The vein lines extracted from Fig. 3 (b)

## V. CONCLUSION

In this paper, we proposed a Retinex based method to reduce the bad influence caused by shadow, and extracted vein lines from palm image. The experiment results show that the proposed method is robust to shadow effect and can better enhance the palm vein image (the contrast ratio increases by 18.4%, entropy increases by 1.07%, and definition increases by 18.8%).

ACKNOWLEDGMENT

We would like to thank for the support from the National Natural Science Foundation of China (Grant No.61301297), the Natural Science Foundation Project of CQ CSTC (No. cstc2012jjA40063), and the Fundamental Research Funds for the Central Universities (No. XDJK2013A006, and XDJK2013C124)